\documentclass[conference]{IEEEtran}
\IEEEoverridecommandlockouts
\usepackage{cite}
\usepackage{amsmath,amssymb,amsfonts}
\usepackage{algorithmicx}
\usepackage{algpseudocode}
\usepackage{algorithm}
\usepackage{algorithm}
\usepackage{graphicx}
\usepackage{epstopdf}
\usepackage{textcomp}
\usepackage{xcolor}
\usepackage{svg} 
\usepackage{booktabs}
\usepackage[table]{xcolor}
\usepackage{multirow}
\usepackage{threeparttable}
\usepackage{subcaption}
\usepackage{hyperref}
\usepackage[section]{placeins}
\def\BibTeX{{\rm B\kern-.05em{\sc i\kern-.025em b}\kern-.08em
    T\kern-.1667em\lower.7ex\hbox{E}\kern-.125emX}}
\begin{document}

\title{Diffusion Posterior Sampling for Super-Resolution under Gaussian Measurement Noise}

\author{\IEEEauthorblockN{Abu Hanif Muhammad Syarubany (20258259)}
\IEEEauthorblockA{
\textit{Korea Advanced Institute of Science \& Technology (KAIST)}\\
Daejeon, South Korea \\
hanif.syarubany@kaist.ac.kr}}

\maketitle

\begin{abstract}
This report studies diffusion posterior sampling (DPS) for single-image super-resolution (SISR) under a known degradation model. We implement a likelihood-guided sampling procedure that combines an unconditional diffusion prior with gradient-based conditioning to enforce measurement consistency for $4\times$ super-resolution with additive Gaussian noise. We evaluate posterior sampling (PS) conditioning across guidance scales and noise levels, using PSNR and SSIM as fidelity metrics and a combined selection score $(\mathrm{PSNR}/40)+\mathrm{SSIM}$. Our ablation shows that moderate guidance improves reconstruction quality, with the best configuration achieved at PS scale $0.95$ and noise standard deviation $\sigma=0.01$ (score $1.45231$). Qualitative results confirm that the selected PS setting restores sharper edges and more coherent facial details compared to the downsampled inputs, while alternative conditioning strategies (e.g., MCG and PS-annealed) exhibit different texture–fidelity trade-offs. These findings highlight the importance of balancing diffusion priors and measurement-gradient strength to obtain stable, high-quality reconstructions without retraining the diffusion model for each operator.
\end{abstract}


\section{Introduction}
Single-image super-resolution (SISR) can be formulated as an inverse problem where a low-resolution observation $y$ is produced from an unknown high-resolution image $x$ via a degradation operator $A$ (e.g., blur+downsampling) and noise $n$, i.e., $y = A(x) + n$. Classical deep SISR approaches learn a direct regression from $y$ to $x$, starting from early CNN baselines such as SRCNN~\cite{Dong2016SRCNN} and progressing to stronger residual architectures such as EDSR~\cite{Lim2017EDSR}. Perceptual objectives and adversarial learning (e.g., SRGAN) further improved visual realism but may hallucinate textures or trade fidelity for perceptual quality~\cite{Ledig2017SRGAN}. Recently, diffusion/score-based generative models have emerged as powerful data priors that enable high-quality synthesis and restoration by iteratively denoising from noise~\cite{Ho2020DDPM,Song2021ScoreSDE}.

In this work, we implement diffusion posterior sampling (DPS) for SISR, leveraging a pretrained unconditional diffusion prior while enforcing measurement consistency through a likelihood-guided update during sampling~\cite{Chung2023DPS}. Concretely, at each reverse diffusion step, we form a denoised estimate $\hat{x}_0$ and update the current sample using the gradient of a data-mismatch objective under the forward model $A(\cdot)$ (Gaussian or Poisson noise), then continue the reverse diffusion process. This yields a simple plug-in framework for inverse problems that does not require retraining the diffusion model for each operator, and is closely related to diffusion-based restoration methods that incorporate measurement operators at inference time~\cite{Kawar2022DDRM,Saharia2023SR3}.

\section{Related Work}
Deep-learning SISR has evolved from early convolutional regressors to very deep residual networks and perceptual models. SRCNN pioneered end-to-end CNN super-resolution and established deep learning as a strong alternative to classical interpolation-based pipelines~\cite{Dong2016SRCNN}. EDSR later demonstrated that carefully-designed residual networks can substantially improve PSNR/SSIM under standard benchmarks~\cite{Lim2017EDSR}. Beyond distortion-oriented objectives, SRGAN introduced adversarial and perceptual losses to enhance visual sharpness and texture realism, albeit sometimes at the cost of strict fidelity to the ground truth~\cite{Ledig2017SRGAN}.

Diffusion/score-based generative models provide a complementary perspective: instead of mapping $y \!\mapsto\! x$ in a single shot, they define iterative stochastic refinement processes that can serve as strong priors~\cite{Ho2020DDPM,Song2021ScoreSDE}. For super-resolution, SR3 (iterative refinement) demonstrated that diffusion models can generate high-quality HR images conditioned on LR inputs~\cite{Saharia2023SR3}. For general image restoration and inverse problems, several works incorporate the forward operator into the sampling procedure; DDRM addresses a family of linear inverse problems via diffusion-based restoration~\cite{Kawar2022DDRM}, while DPS provides a principled and practical likelihood-guided sampling rule that supports noisy and potentially nonlinear operators without retraining~\cite{Chung2023DPS}. Our implementation follows the DPS paradigm for SISR by integrating measurement-gradient guidance into the reverse diffusion loop.

\section{Methodology}
\label{sec:method}

\subsection{Inverse Problem Formulation (Super-Resolution)}
Given a low-resolution (LR) measurement $y \in \mathbb{R}^{m}$, we aim to recover
a high-resolution (HR) image $x_0 \in \mathbb{R}^{d}$ through the (possibly noisy)
forward model
\begin{equation}
y = \mathcal{A}(x_0) + n,
\label{eq:forward_model}
\end{equation}
where $\mathcal{A}:\mathbb{R}^{d}\rightarrow\mathbb{R}^{m}$ denotes the degradation
operator (e.g., downsampling for super-resolution) and $n$ is measurement noise.
Our goal is to sample from the posterior $p_\theta(x_0\mid y)$ using a diffusion prior
$p_\theta(x_0)$ learned from clean HR images.

\subsection{Diffusion Prior and Unconditional Reverse Sampling}
We adopt a denoising diffusion model as a powerful image prior. The forward diffusion
process gradually perturbs a clean image $x_0$ into noisy latents $x_t$:
\begin{equation}
q(x_t\mid x_0) = \mathcal{N}\!\left(x_t; \sqrt{\bar{\alpha}_t}\,x_0,\; (1-\bar{\alpha}_t)\mathbf{I}\right),
\label{eq:forward_diffusion}
\end{equation}
where $\bar{\alpha}_t=\prod_{s=1}^{t}\alpha_s$ and $\alpha_s=1-\beta_s$ with a predefined
noise schedule $\{\beta_t\}_{t=1}^{T}$ (linear or cosine in our implementation).
Sampling starts from $x_T\sim \mathcal{N}(0,\mathbf{I})$ and proceeds via the learned
reverse process $p_\theta(x_{t-1}\mid x_t)$.

\begin{figure}[t]
    \centering
    \includegraphics[width=\columnwidth]{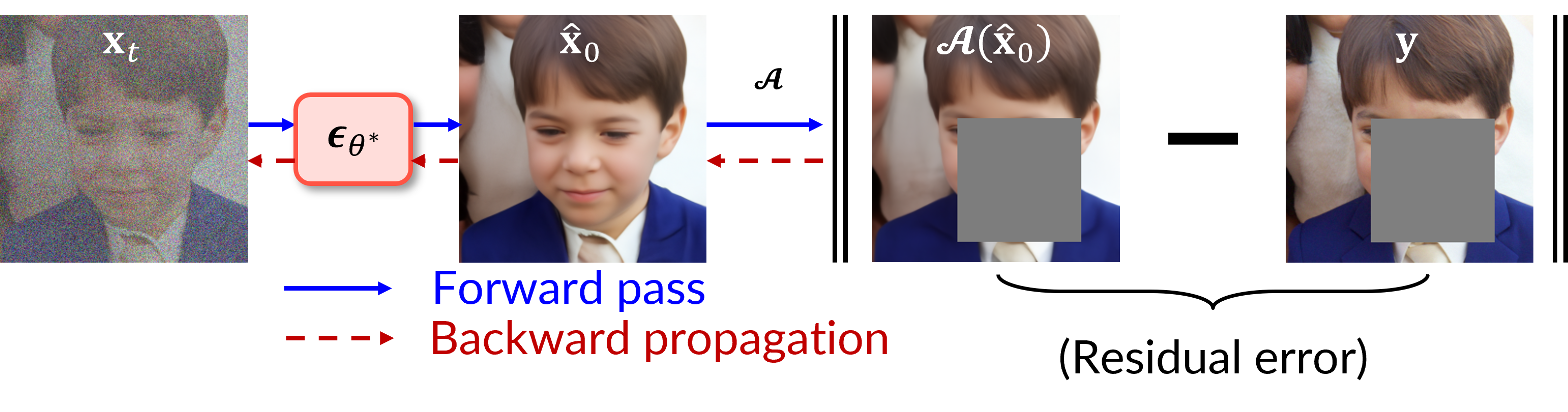} 
    \caption{DPS guidance pipeline at timestep $t$. The denoiser predicts $\hat{\epsilon}_\theta(x_t,t)$ and forms $\hat{x}_0(x_t,t)$ via Tweedie’s formula. The forward operator produces $\mathcal{A}(\hat{x}_0)$, which is compared to the measurement $y$ to compute a residual. Gradients are backpropagated through $\mathcal{A}(\cdot)$ and $\hat{x}_0(x_t,t)$ to update the sample for measurement-consistent reverse diffusion.}
    \label{fig:dps_pipeline}
\end{figure}

In practice, the network predicts a denoising direction (or noise) that is converted
into an estimate of the clean image $\hat{x}_0$ at time $t$. Consistent with the
pipeline shown in Fig.~\ref{fig:dps_pipeline}, we use the standard Tweedie-form estimator
\begin{equation}
\hat{x}_0(x_t,t)
=
\frac{1}{\sqrt{\bar{\alpha}_t}}
\left(
x_t - \sqrt{1-\bar{\alpha}_t}\;\hat{\epsilon}_\theta(x_t,t)
\right),
\label{eq:tweedie}
\end{equation}
which corresponds to the \texttt{pred\_xstart} returned by our sampler.

\subsection{Conditional Sampling via Denoising Posterior Sampling (DPS)}
Direct conditioning requires the likelihood term at diffusion time $t$:
\begin{equation}
\nabla_{x_t}\log p_t(x_t\mid y)
=
\nabla_{x_t}\log p_t(x_t)
+
\nabla_{x_t}\log p_t(y\mid x_t),
\label{eq:guidance_decomp}
\end{equation}
but $p_t(y\mid x_t)=\int p(y\mid x_0)p(x_0\mid x_t)\,dx_0$ is intractable.

DPS approximates this integral using Jensen's inequality by replacing the random
variable $x_0\sim p(x_0\mid x_t)$ with its conditional mean estimate $\hat{x}_0(x_t,t)$:
\begin{equation}
p_t(y\mid x_t)
=\mathbb{E}_{p(x_0\mid x_t)}[p(y\mid x_0)]
\;\approx\;
p\big(y\mid \hat{x}_0(x_t,t)\big).
\label{eq:dps_jensen}
\end{equation}
For Gaussian measurements, we use
$p(y\mid x_0)=\mathcal{N}(y;\mathcal{A}(x_0),\sigma^2\mathbf{I})$.
Thus, DPS guidance becomes a gradient step that reduces the data mismatch:
\begin{equation}
x_t \leftarrow x_t - \rho\;\nabla_{x_t}\,\mathcal{L}\!\left(y,\mathcal{A}(\hat{x}_0)\right),
\qquad
\mathcal{L}(\cdot)=\left\lVert y-\mathcal{A}(\hat{x}_0)\right\rVert_2,
\label{eq:dps_update}
\end{equation}
where $\rho>0$ is the guidance scale. (We use the $\ell_2$ norm, matching the code path
\texttt{torch.linalg.norm}; a Poisson option is implemented with a normalized residual
surrogate.)

\paragraph{Backpropagation through the pipeline.}
Following Fig.~\ref{fig:dps_pipeline}, the residual is computed in measurement space,
then backpropagated through $\mathcal{A}(\cdot)$ and the Tweedie estimator
$\hat{x}_0(x_t,t)$ to obtain the gradient w.r.t.\ the current latent. In our implementation,
we enable autograd on the current iterate (named \texttt{x\_prev} in code), compute
$\hat{x}_0$ via the denoiser forward pass, and apply the guidance update to the newly
sampled latent at the same timestep.

\subsection{Conditioning Modules Used in Our Implementation}
We implement conditioning as a modular interface (\texttt{ConditioningMethod}) with:
(i) a forward operator \texttt{operator.forward} implementing $\mathcal{A}(\cdot)$,
(ii) an optional projection \texttt{operator.project}, and (iii) a noise model flag
(\texttt{gaussian} or \texttt{poisson}).
We support the following conditioning methods:
\begin{itemize}
    \item \textbf{Vanilla (Identity):} unconditional sampling without measurement guidance.
    \item \textbf{Projection:} projects the iterate to satisfy measurement consistency using a
    time-matched noisy measurement $\tilde{y}_t\sim q(\cdot\mid y)$ (Eq.~\eqref{eq:forward_diffusion}).
    \item \textbf{Posterior Sampling (DPS, \texttt{ps}):} applies the gradient step
    in Eq.~\eqref{eq:dps_update} using scale $\rho$ (\texttt{scale}).
    \item \textbf{Manifold Constraint Gradient (MCG, \texttt{mcg}):} combines
    a DPS-like gradient step with a projection step, improving stability when strict
    measurement consistency is desired.
\end{itemize}


\begin{algorithm}[t]
\caption{ DPS-Guided Diffusion Sampling (Super-Resolution)}
\label{alg:dps_sampling}
\begin{algorithmic}[1]

\Require Diffusion model $\hat{\epsilon}_\theta$, operator $\mathcal{A}$, measurement $y$, steps $T$, guidance scale $\rho$
\Ensure HR reconstruction sample $x_0$
\State Initialize $x_T \sim \mathcal{N}(0,\mathbf{I})$
\For{$t=T-1,\dots,0$}
    \State Set $x_t \leftarrow x_t$ with \texttt{requires\_grad=True}
    \State Unconditional step: $(x_{t-1}^{\text{unc}},\hat{x}_0)\leftarrow p_\theta(x_{t-1}\mid x_t)$
    \State DPS loss: $\mathcal{L}\leftarrow \|y-\mathcal{A}(\hat{x}_0)\|_2$
    \State Gradient: $g \leftarrow \nabla_{x_t}\mathcal{L}$ \Comment{via backprop through Fig.~\ref{fig:dps_pipeline}}
    \State Guided update: $x_{t-1} \leftarrow x_{t-1}^{\text{unc}} - \rho\, g$
    \State Detach $x_{t-1}$ (free graph/memory)
\EndFor
\State \Return $x_0 \leftarrow x_{0}$
\end{algorithmic}
\end{algorithm}

\begin{algorithm}[t]
\caption{Conditioning Step (PS / MCG) in Our Code}
\label{alg:conditioning}
\begin{algorithmic}[1]
\Require Current latent $x_{\text{prev}}$, proposed latent $x_t$, denoised estimate $\hat{x}_0$, measurement $y$, scale $\rho$
\Ensure Conditioned latent $\tilde{x}_t$ and residual value
\State Residual $r \leftarrow y - \mathcal{A}(\hat{x}_0)$
\If{Gaussian}
    \State $\mathcal{L}\leftarrow \|r\|_2$
\ElsIf{Poisson (surrogate)}
    \State $\mathcal{L}\leftarrow \text{mean}\!\left(\|r\|_2 / (|y|+\varepsilon)\right)$
\EndIf
\State $g \leftarrow \nabla_{x_{\text{prev}}}\mathcal{L}$ \Comment{ autograd in \texttt{grad\_and\_value}}
\State $x_t \leftarrow x_t - \rho\, g$ \Comment{ \texttt{PosteriorSampling} update}
\If{MCG}
    \State Draw time-matched noisy measurement $\tilde{y}_t \sim q(\cdot\mid y)$
    \State $x_t \leftarrow \texttt{project}(x_t,\tilde{y}_t)$ \Comment{ \texttt{operator.project}}
\EndIf
\State \Return $(\tilde{x}_t=x_t,\;\mathcal{L})$
\end{algorithmic}
\end{algorithm}

\subsection{Practical Notes (Implementation Mapping)}
Our implementation directly mirrors the above methodology:
(i) the reverse sampler (\texttt{DDPM} or \texttt{DDIM}) produces an unconditional proposal
and $\hat{x}_0$ (\texttt{pred\_xstart}); (ii) DPS/MCG computes the measurement residual
$\|y-\mathcal{A}(\hat{x}_0)\|_2$ and backpropagates to obtain a guidance direction; and
(iii) the iterate is updated and optionally projected using a time-noised measurement
$\tilde{y}_t$ to remain consistent with the diffusion noise level.

\section{Experiments and Results}
\label{sec:exp_results}

\subsection{Experimental Setup}
We evaluate \emph{4$\times$ single-image super-resolution} under the measurement model
$y = A(x) + n$, where $A(\cdot)$ is the super-resolution degradation operator (downsampling by a factor of $4$) and
$n \sim \mathcal{N}(0,\sigma^2)$ is additive Gaussian noise applied to the measurement.
For sampling, we use a diffusion model with a U-Net backbone and perform guided reconstruction via \textbf{Posterior Sampling (PS)} conditioning, which updates the current sample in the direction that reduces the measurement residual.

Unless stated otherwise, we use the following diffusion configuration:
\begin{verbatim}
sampler: ddpm
steps: 1000
noise_schedule: linear
model_mean_type: epsilon 
model_var_type: learned_range
dynamic_threshold: False
clip_denoised: True
rescale_timesteps: False
timestep_respacing: 1000
\end{verbatim}

\subsection{Quantitative Results}
We report reconstruction quality using PSNR and SSIM, and for configuration selection we additionally track the combined score
\begin{equation}
\label{eq:combined_score}
\text{Score} \;=\; \frac{\text{PSNR}}{40} + \text{SSIM},
\end{equation}
where $40$ (dB) is used as a normalization constant for PSNR. Table~\ref{tab:ps_ablation_scale_sigma} summarizes the effect of the PS guidance scale and the Gaussian noise level $\sigma$ on the combined score.
Overall, the best configuration (scale $=0.95$, $\sigma=0.01$) suggests that \emph{slightly weaker guidance} than the default and a \emph{cleaner measurement target} produce a better prior--likelihood balance.
At a fixed scale ($0.8$), decreasing $\sigma$ from $0.07 \rightarrow 0.005$ consistently improves the score, indicating that measurement noise directly degrades both fidelity (PSNR) and structure preservation (SSIM) under PS guidance.
Conversely, overly small scales (e.g., $0.5$ and $0.2$ at $\sigma=0.05$) substantially reduce the score, consistent with under-enforcing measurement consistency and yielding less effective super-resolution.

\begin{table}[t]
\centering
\caption{PS conditioning ablation on guidance scale and Gaussian noise level.}
\label{tab:ps_ablation_scale_sigma}
\begin{threeparttable}
\setlength{\tabcolsep}{6pt}
\begin{tabular}{c c c}
\toprule
\textbf{PS scale} & \textbf{Noise $\sigma$} & \textbf{$(\mathrm{PSNR}/40)+\mathrm{SSIM}$} \\
\midrule
0.95 & 0.01  & \textbf{1.45231} \\
0.9 & 0.05  & 1.40065 \\
0.9 & 0.01  & 1.44452 \\
0.8 & 0.07  & 1.36506 \\
0.8 & 0.05  & 1.38599 \\
0.8 & 0.03  & 1.40976 \\
0.8 & 0.01  & 1.42857 \\
0.8 & 0.005 & 1.42954 \\
0.5 & 0.05  & 1.32122 \\
0.2 & 0.05  & 1.16456 \\
\bottomrule
\end{tabular}
\begin{tablenotes}[flushleft]
\footnotesize
\item \textit{Remark:} ``PS scale'' controls the strength of the gradient-based guidance step (larger $\Rightarrow$ stronger enforcement of measurement consistency).
``Noise $\sigma$'' is the standard deviation of the Gaussian noise injected into the measurement $y$.
\end{tablenotes}
\end{threeparttable}
\end{table}

\subsection{Configuration Selection}
Based on Table~\ref{tab:ps_ablation_scale_sigma}, we select \textbf{PS conditioning} with \textbf{scale $=0.95$} and \textbf{noise $\sigma=0.01$}, which achieves the best combined score.
Empirically, too small a PS scale under-enforces measurement consistency (leading to weaker super-resolution), while too large a scale can over-correct each step and introduce instability or artifacts.
Likewise, smaller $\sigma$ yields a cleaner measurement target and improves the consistency loss signal, but overly small values may reduce robustness when the measurement operator or data distribution is imperfect.

\subsection{Qualitative Results}
\begin{figure}[t]
\centering
\includegraphics[width=\linewidth]{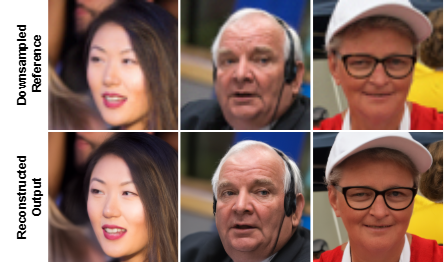}
\caption{Qualitative comparison for super-resolution using the selected PS configuration (scale $=0.9$, $\sigma=0.01$). Top: downsampled measurement (input). Bottom: reconstructed output.}
\label{fig:qual_ps_best}
\end{figure}

\begin{figure}[t]
\centering
\includegraphics[width=\linewidth]{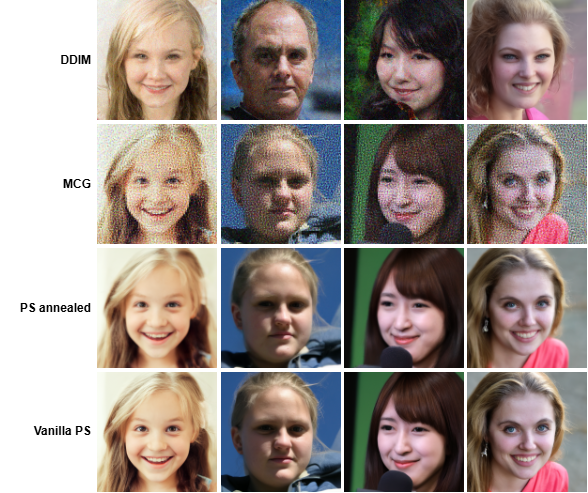}
\caption{Qualitative comparison across samplers/conditioning methods. First row: DDIM sampler. Remaining rows: DDPM sampler with different conditioning methods (MCG, PS annealed, Vanilla PS).}
\label{fig:qual_method_comp}
\end{figure}

Figure~\ref{fig:qual_ps_best} visualizes the qualitative improvement from the chosen PS configuration, comparing the downsampled input (measurement) to the reconstructed high-resolution output.
In addition to sharper edges, the PS output better restores mid-frequency facial details (e.g., contours around cheeks/mouth) while avoiding excessive ringing, indicating that the diffusion prior still dominates when the measurement gradient becomes unreliable.
Figure~\ref{fig:qual_method_comp} compares multiple conditioning/sampling strategies; the first row uses a DDIM sampler, while the remaining rows use DDPM. Visually, MCG tends to introduce stronger texture-like artifacts and noise amplification, whereas PS-annealed and vanilla PS better preserve smooth regions (skin) with more coherent high-frequency details.
We also observe that DDIM produces different texture characteristics (often smoother but sometimes less detailed) due to its more deterministic trajectory, while DDPM-based PS variants yield more natural stochastic detail when the guidance is appropriately scaled.

\subsection{Discussion}
\textbf{Why smaller PS scale can improve PSNR/SSIM.}
In PS conditioning, the update typically follows a negative gradient step that reduces the measurement residual (e.g., $\|A(\hat{x}_0)-y\|$).
A large scale may overshoot the descent direction at each diffusion step, causing oscillations and visually implausible corrections, which can degrade perceptual quality and structural similarity.
Reducing the scale (e.g., $1.0 \rightarrow 0.9$) often yields a better balance between (i) the diffusion prior (natural image manifold) and (ii) measurement consistency, improving PSNR/SSIM.

\textbf{PS vs. MCG vs. PS-annealed.}
PS (Posterior Sampling / DPS-style guidance) directly nudges samples toward measurement consistency via gradient guidance.
MCG additionally enforces a hard projection onto the measurement manifold, which can amplify artifacts when the projection is aggressive or when the operator/noise assumptions are imperfect.
PS-annealed varies the guidance strength over timesteps (typically weaker early, stronger late), aiming to preserve global structure early while enforcing consistency near the end; this can improve stability but may require careful scheduling to avoid under- or over-guidance.
In our qualitative comparisons, PS (and PS-annealed) provides a strong trade-off between sharpness and naturalness under DDPM sampling.

\section{Conclusion}
We implemented a diffusion posterior sampling framework for $4\times$ SISR by integrating a super-resolution measurement operator into the reverse diffusion process through gradient-based conditioning. Quantitatively, a simple ablation over PS scale and Gaussian noise level shows that moderate guidance yields the best performance, with scale $0.95$ and $\sigma=0.01$ achieving the highest combined score $(\mathrm{PSNR}/40)+\mathrm{SSIM}=1.45231$. Qualitatively, the selected configuration produces sharper and more natural reconstructions than the downsampled measurements, while method comparisons suggest that stronger constraint mechanisms (e.g., MCG) can amplify artifacts if the projection or noise assumptions are mismatched.

\bibliographystyle{IEEEtran}
\bibliography{myblib}
\end{document}